\documentclass[3p,times]{elsarticle}

\usepackage{ecrc}

\volume{00}

\firstpage{1}

\journalname{Neural Networks}

\runauth{}

\jid{procs}

\jnltitlelogo{Neural Networks}

\CopyrightLine{2011}{Published by Elsevier Ltd.}

\usepackage{amssymb}

\usepackage[figuresright]{rotating}

\usepackage{multirow}
\usepackage{makecell}
\usepackage{xcolor}
\usepackage{amsfonts}
\usepackage{comment}

\newcommand{\who}[2]{\cite{#2}}
\newcommand{\ie}{\textit{i}.\textit{e}., }
\newcommand{\eg}{\textit{e}.\textit{g}., }
\renewcommand*\cite[1]{\citep{#1}}

\begin{document}

\begin{frontmatter}



\dochead{}

\title{Content Preserving Image Translation with Texture Co-occurrence and Spatial Self-Similarity for Texture Debiasing and Domain Adaptation}


\author[1]{Myeongkyun~Kang}
\author[1]{Dongkyu~Won}
\author[1]{Miguel~Luna}
\author[1]{Philip~Chikontwe}
\author[2]{Kyung~Soo~Hong}
\author[2]{June~Hong~Ahn}
\author[1]{Sang~Hyun~Park\corref{cor1}}
\ead{shpark13135@dgist.ac.kr}
\cortext[cor1]{Corresponding author.}

\address[1]{Department of Robotics and Mechatronics Engineering, Daegu Gyeongbuk Institute of Science and Technology (DGIST), Daegu, South Korea.}
\address[2]{Division of Pulmonology and Allergy, Department of Internal Medicine, Regional Center for Respiratory Diseases, Yeungnam University Medical Center, College of Medicine, Yeungnam University, Daegu, South Korea.}

\begin{abstract}
Models trained on datasets with texture bias usually perform poorly on out-of-distribution samples since biased representations are embedded into the model. Recently, various image translation and debiasing methods have attempted to disentangle texture biased representations for downstream tasks, but accurately discarding biased features without altering other relevant information is still  challenging. In this paper, we propose a novel framework that leverages image translation to generate additional training images using the content of a source image and the texture of a target image with a different bias property to explicitly mitigate texture bias when training a model on a target task. Our model ensures texture similarity between the target and generated images via a texture co-occurrence loss while preserving content details from source images with a spatial self-similarity loss. Both the generated and original training images are combined to train improved classification or segmentation models robust to inconsistent texture bias. Evaluation on five classification- and two segmentation-datasets with known texture biases demonstrates the utility of our method, and reports significant improvements over recent state-of-the-art methods in all cases.
\end{abstract}


\end{frontmatter}


\section{Introduction}\label{sec:introduction}

\begin{figure}[!t]
	\centering
	\includegraphics[width=0.5\columnwidth]{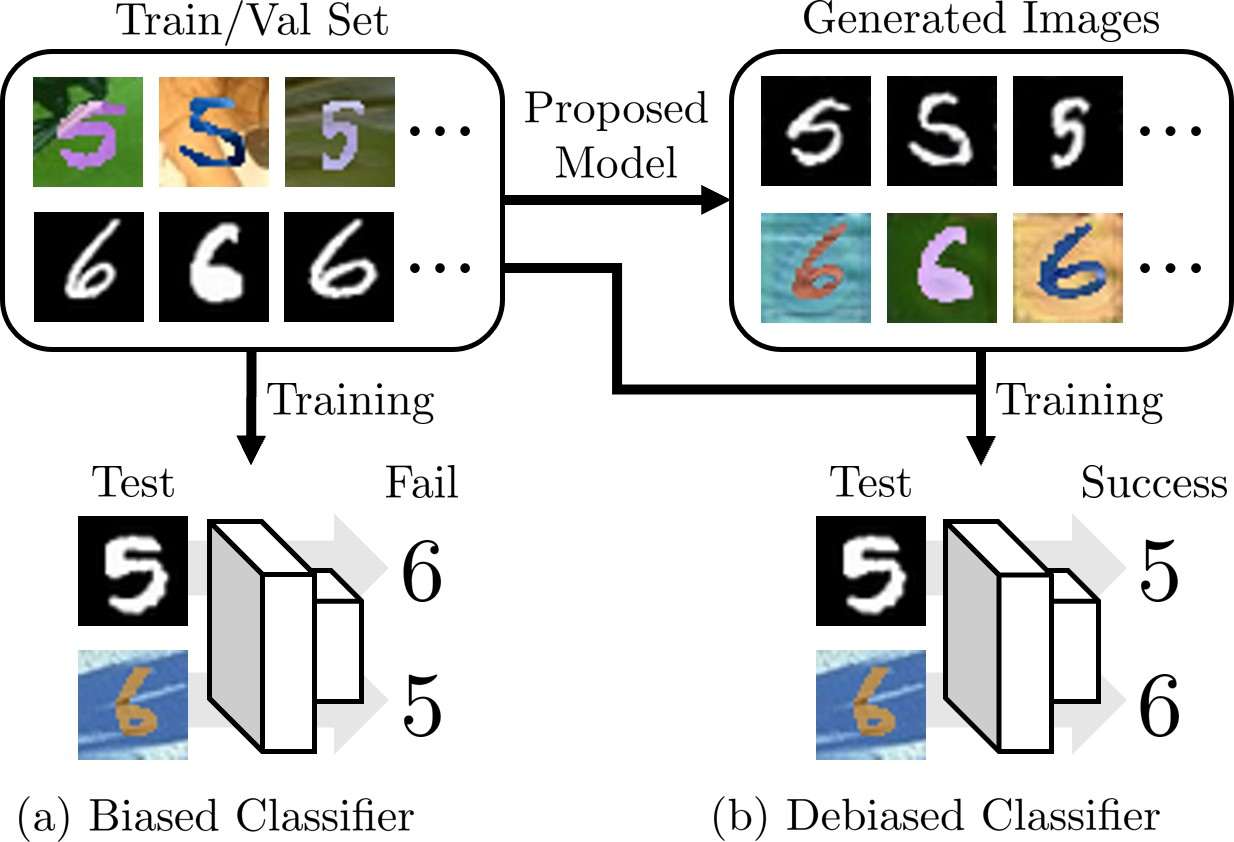}
	\caption{(a) A classifier learned using texture biased data often gives wrong predictions. (b) We extend the training dataset with texture augmented images to train a debiased classifier.}
	\label{fig_concept}
\end{figure}

Texture biases can be easily and unintentionally introduced during the data collection process. The source and properties of such biases are usually unknown, and can lead to significant drops in performance when a model trained on biased data is applied to out-of-distribution data \cite{barbu2019objectnet,bissoto2020debiasing}. For instance, \who{Geirhos}{geirhos2018imagenet} noted that common convolutional neural networks (CNNs) prioritize texture information over content features (\eg shape); making them vulnerable, especially on images with texture bias. As shown in Fig. \ref{fig_concept}(a), if a binary classifier is trained with images that have a distinct texture for each class (\eg five has color and six is grayscale), it is highly likely that the model will consider texture rather than the actual shape for classification. Consequently, the model will perform poorly on test data that does not include similar textures as those observed during training. Therefore, transferring texture features between images of different classes allows a classifier to adjust its internal learned representation to be less dependent on bias information.

Bias mitigation or debiasing is a well-studied problem in literature, prior works proposed to extract bias-independent features through adversarial learning and enabled models to solve the intended classification task without relying on biased features (\eg color information) \cite{kim2019learning,wang2019balanced}. However, texture biased representations are often retained after training since it is difficult to completely disentangle biased features through adversarial learning. Also, prior knowledge regarding the types of biases in the training data is essential for adversarial learning, but is often unknown and difficult to identify, especially when the data has been collected from multiple sources \cite{barbu2019objectnet,bissoto2020debiasing,goyal2017making,li2018resound}.

Along this line of thought, similar scenarios may arise when curating a segmentation dataset under some specific setting \eg using graphics game engines or machines, resulting in lower performance during testing due to texture discrepancies arising from data acquisition. In this scenario, unsupervised domain adaptation (UDA) methods have been proposed to handle texture discrepancies among domains via alignment at the image-, feature-, output-levels as well as using self-training approaches \cite{hoffman2018cycada,tsai2018learning,li2019bidirectional,zou2018unsupervised,mei2020instance,kim2020learning,tranheden2021dacs,luo2023adversarial,li2022cross,hou2022deep,wang2022informative}.
However, these approaches are highly rely on complicate design choices, and the texture discrepancies among domains can be easily resolved when the image translation performs well enough.  

In this paper, we propose an efficient and practical strategy where an image translation model is used to facilitate improved downstream performance.
We consider employing image translation as a reasonable alternative over prior complex designs, applicable to both debiasing and domain adaptation.
We extend a given dataset with additional generated/translated unbiased images; so as to alleviate learning biased representations when training for downstream tasks.
Specifically, we combine the structural information of a source image with the texture of another labeled image to extend the dataset. For debiasing, the extended dataset helps the classifier learn class related features and avoid relying on biased features to perform correct data separation. As shown in Fig. \ref{fig_concept}(b), a debiased classifier can be trained without bias labels by combining the original data and generated images containing the textures of images with different labels. For domain adaptation, the extended dataset helps the segmentation model learn the shifted training distribution to perform well on real-world samples or other biased samples.

Modern image translation methods \cite{zhu2017unpaired,liu2017unit,huang2018munit,lee2020drit++,kolkin2019style,park2020cut,sun2022face,zheng2022not,lei2023lac} have shown remarkable success in effectively transferring textures (or styles) between pairs of images, but they also tend to adjust content information since no explicit supervision is provided during training \cite{hoffman2018cycada,cohen2018distribution}. For instance, in Fig. \ref{fig_concept}, the ideal scenario would be the colored digit five is translated to black and white, but is instead translated into a six that matches the black and white style. In general, transferring only texture information without content discrepancy is very challenging, and poor results will adversely affect downstream tasks.
Thus, additional constraints are required to ensure the content retainment when the strong texture loss enforce to generate target-like images.

To address this issue, we propose a novel image translation method that simultaneously considers spatial self-similarity between a source image and its generated image, and texture similarity between the generated image and a target image having a different label than that of the source image. The proposed framework consists of an image generator with content and texture encoders, two discriminators that constrain texture similarity in both local and global views, and a pre-trained VGG \cite{simonyan2014very} network to enforce spatial self-similarity. The generation of high quality images with the intended properties is achieved by using a \textit{spatial self-similarity loss} to ensure content consistency, \textit{texture co-occurrence} and \textit{GAN losses} to enforce similar local and global textures in the target image. Once images are generated using the training data, a debiased classifier or adapted segmentation model can be learned using all available data. The main contributions are listed below:

\begin{itemize}

\item 
We propose a novel image translation method employing texture co-occurrence and spatial self-similarity losses. While these losses have been respectively proposed to address different tasks, both have never been considered jointly for the image translation task. We show that optimizing both losses can produce images that are effective for both debiasing and domain adaptation.

\item 
We introduce a method to learn a debiased classifier by explicitly augmenting data rather than designing a complex model. Our method does not require any bias labels and can effectively mitigate unknown biases during training.

\item
We further employ our method to adapt segmentation to the target domain by explicitly augmenting the training dataset. Our method is fully decoupled to the segmentation module and can adapt newly proposed segmentation model more tractable.

\item
We show that our method outperforms existing debiasing and domain adaptation methods, and can produce high quality images compared to prior image translation models on five biased datasets and two domain adaptation datasets, respectively.

\end{itemize}

This paper is organized as follows. Section \ref{sec:relatedworks} presents the related works. Section \ref{sec:method} presents our proposed method. 
The dataset and experimental details are given in Section \ref{sec:experiments}. The results are shown in Section \ref{sec:results}.
Finally, Section \ref{sec:conclusion} concludes the paper.

\section{Related Works}\label{sec:relatedworks}

\subsection{Image Translation and Style Transfer}
In literature, image translation methods often employ cycle-consistency and contrastive losses to preserve content during training \cite{zhu2017unpaired,liu2017unit,huang2018munit,lee2020drit++,park2020cut,zheng2021spatially} while style transfer methods employ the perceptual loss \cite{gatys2016image} and it's variants for optimization \cite{kolkin2019style}.
\who{Zhu}{zhu2017unpaired} and \who{Liu}{liu2017unit} proposed deep generative models using a pair of generators and discriminators that translate one domain into another using cycle-consistency and a shared-latent space assumption, respectively. \who{Huang}{huang2018munit} proposed a multi-domain translation model by varying the style code with fixed content for diverse style image generation. For feature disentanglement, \who{Lee}{lee2020drit++} proposed a content and attribute encoder with a cross-cycle consistency loss to enforce consistency between domains. Furthermore, \who{Park}{park2020cut} proposed a conditional image translation method with a contrastive loss to maximize the mutual information between positive and negative patches, whereas \who{Zheng}{zheng2021spatially} proposed a spatially-correlative loss for consistent image translation to preserve scene structures.
Meanwhile, \who{Gatys}{gatys2016image} proposed an optimization-based style transfer method using the Gram-matrix with a pre-trained VGG model, and \who{Kolkin}{kolkin2019style} designed a content preservation variant using the concept of self-similarity.
Recently, \who{Park}{park2020swapping} proposed a method that enforces co-occurrent patch statistics across different parts of the image for image manipulation.
Note that as most GAN-based methods focused on transferring image style from one domain to another instead of maintaining image content; spatial discrepancy or texture corruption may be observed in the generated images.
Moreover, since semantic changes and pixel mismatches adversely affect downstream tasks - poor results are expected; thus, we believe preserving  image content is important.
Unlike prior generation models, our model only updates textures while preserving content by minimizing a spatial self-similarity loss \cite{kolkin2019style} whereas a texture loss enforces the generation of target-like images.

\subsection{Texture Bias and Debiasing Methods}
\who{Geirhos}{geirhos2018imagenet} argued that texture bias mitigation is necessary to ensure the reliability of a classifier, since common CNNs prioritize texture information over content (shape). Especially in the task of domain generalization \cite{zhou2021domainsurvey}, texture bias has gained significant interest since changes in image texture are the main reason for domain shift \cite{wang2019learning,nam2021reducing,zhou2021domain,venkateswara2017deep,li2017deeper,peng2019moment}. \who{Nuriel}{nuriel2021permuted} and \who{Zhou}{zhou2021domain} applied feature-level adaptive instance normalization (AdaIN) \cite{huang2017arbitrary} by shuffling (or swapping) the features of training samples across source domains to improve the generalizability of the trained model. Similarly, \who{Nam}{nam2021reducing} introduced content- and style-biased networks that randomize the content and style (texture) features between two different samples via AdaIN. To obtain features robust to style-bias, they also leveraged adversarial learning to prevent the feature extractor from retaining style-biased representations. In general, prior methods only focused on the generalization performance of inaccessible-domain samples, and thus design models to learn common object features from multiple source domains. However, if the sample is biased along the label \eg colored  digit  five and  grayscale digit six in Fig. \ref{fig_concept}, existing methods may fail to learn common object features, and cannot effectively ignore texture biases present in  the training dataset. In this paper, we address this by training a classifier to only focus on the intended task without using bias information present in the training data.

For debiasing, diverse losses and balancing approaches are proposed to mitigate bias and stabilizing training \cite{alvi2018turning,kim2019learning,sattigeri2019fairness,choi2020fair,li2019repair,wang2019balanced,louppe2017learning,zhang2018mitigating}.
To mitigate bias, \who{Alvi}{alvi2018turning} employed a bias prediction layer to make latent features indistinguishable using a confusion loss \cite{tzeng2015simultaneous}. \who{Kim}{kim2019learning} used a gradient reversal layer \cite{ganin2015unsupervised} that minimizes the mutual information of bias predictions to constrain the use of bias-related features for classification.
In addition, few related works \cite{sattigeri2019fairness,choi2020fair} used the fairness term in GANs. These methods were proposed for the generation of a fraction (e.g., 10\%) of protected attribute samples for a model trained on a biased dataset, and could not generate images with a specific texture. We instead employ an image translation technique that explicitly generates an image with the texture of a target image. On the subject of fairness aware training, \who{Li}{li2019repair} formulated bias minimization in terms of data re-sampling to balance the preference of specific representations for classification.
On the other hand, \who{Wang}{wang2019balanced} employed an adversarial learning approach \cite{ganin2015unsupervised} to remove protected attributes correspondence (\eg gender) in the intermediate features of the model. Moreover, \who{Louppe}{louppe2017learning} proposed an adversarial network to enforce the pivotal property (\eg fairness) on a predictive model, and \who{Zhang}{zhang2018mitigating} proposed three terms, \ie primary, adversary, and projection, to improve the stability of debias training. However, these methods require bias labels for training and stability in adversarial learning is often hard to achieve. In contrast, our method solves the bias problem by explicitly utilizing texture generated images and only requires images of different domains. Thus, our approach does not require laborious bias labeling and is free from intractable adversarial classifier training.

\subsection{Unsupervised Domain Adaptation}
Unsupervised domain adaptation (UDA) methods in semantic segmentation have been proposed to address the domain shift problem where source and target domains are known in advance. Most use adversarial learning to reduce the gaps between the source and target domains at the input-, feature- and output-levels, or using self-(semi-) training and data augmentation \cite{hoffman2018cycada,tsai2018learning,li2019bidirectional,zou2018unsupervised,mei2020instance,kim2020learning,tranheden2021dacs}.
\who{Hoffman}{hoffman2018cycada} first translated an image from the source to the target domain using cycle-consistency, then applied adversarial learning on the segmentation backbone features to further reduce domain discrepancies. \who{Tsai}{tsai2018learning} employed adversarial learning on the output of the segmentation model since semantic maps do not vary significantly between domains despite the presence of inconsistencies in pixels.
\who{Li}{li2019bidirectional} translate an image to adapt to the target domain, and also employ a trained segmentation model to enforce content consistency preservation of the image translation model. 

For self-training UDA, \who{Zou}{zou2018unsupervised} used class normalized confidence to generate class-balanced pseudo labels considering hard class predictions. \who{Mei}{mei2020instance} additionally set instance-level class thresholds per image to consider both local and global characteristics for pseudo label generation, with local class thresholds used to update the global class thresholds based on the exponential moving average. Lastly, in the case of data augmentation for UDA, \who{Kim}{kim2020learning} employed style-transfer, and \who{Tranheden}{tranheden2021dacs} used cross-domain mixed samples that crop target images using its pseudo masks and attaching it to source images.
Though successful, note that performance is highly dependent on complex design choices.
We instead directly train a segmentation model using translated images that are intended to preserve content of source images.
This approach is more efficient and practical since the translation quality can be explicitly verified and can be applied a recent state-of-the-art segmentation model in a more tractable way.

\begin{figure}[!t]
	\centering
	\includegraphics[width=0.9\textwidth]{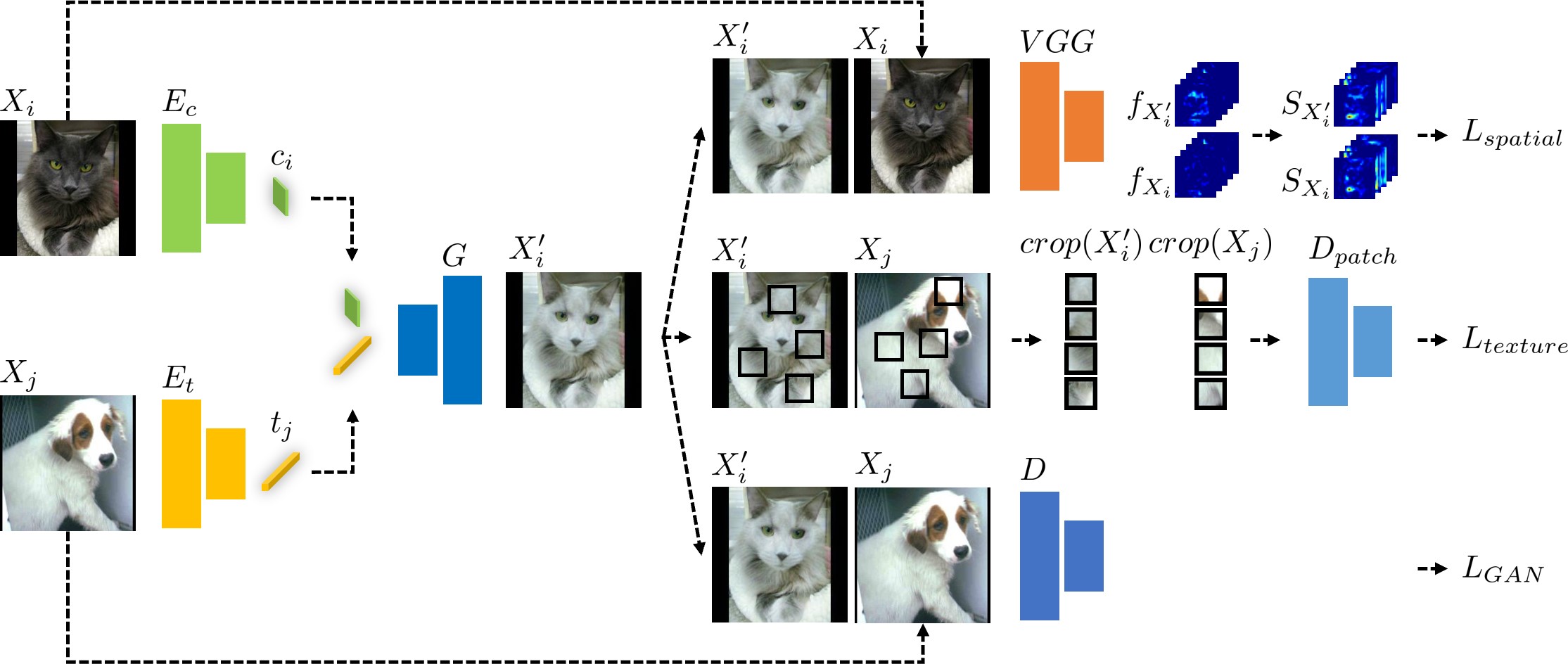}
	\caption{Our proposed image generation model. Our model consists of a content encoder $E_c$, texture encoder $E_t$, StyleGAN2 generator $G$, ImageNet pre-trained $VGG$ model, and two discriminators $D$ and $D_{patch}$. Each encoder extracts the content feature $c_i$ from an image $X_i$ and the texture feature $t_j$ in an image $X_j$ with an different label, respectively. $G$ generates an image $X^{\prime}_i$ using $c_i$ and $t_j$ with a spatial self-similarity loss $L_{spatial}$, a texture co-occurrence loss $L_{texture}$, and a GAN loss $L_{GAN}$.}
	\label{fig_model}
\end{figure}

\section{Method}\label{sec:method}
Given an image dataset $\mathcal{D} = \{X_1,...,X_n\}$ with binary bias (or domain) labels $b_i \in \{0,1\}$, and classification labels $y_i \in \{0,1\}$ aligned with the bias property \ie $b_i=y_i$ for a biased classification dataset. Our main goal is to build an augmented dataset $\mathcal{D}^{\prime}=\{X^{\prime}_{1},...,X^{\prime}_{n}\}$ that reduces (or adapts) the importance of $b_i$ on the target task process. In this paper, we consider texture as the dominant property that creates bias during target task learning, thus discourages the model from using the shape information of the object of interest, as well as aligning the training distribution for debiasing and adaptation, respectively. Therefore, we propose a generative data augmentation framework that updates an input image $X_i$ using the texture of a randomly selected image with a different bias property \ie $X_{j, b_i \neq b_j}$, while retaining the content information of $X_i$.

Our framework consists of three main sections as shown in Figure \ref{fig_model}. First, $X_i$ and $X_j$ are encoded by a content encoder $E_c$ and a texture encoder $E_t$, respectively. Then, the encoded features are used to generate a target augmented image $X^{\prime}_i$ via $G$, and a combination of two discriminators with an ImageNet pre-trained $VGG$ model \cite{simonyan2014very} ensure the generated image $X^{\prime}_i$ has the texture of $X_j$ while retaining the content information of $X_i$. Modifying texture information while maintaining the content requires the use of additional terms in the standard adversarial objective. Thus, we add a texture co-occurrence loss term to enable correct texture transfer and a spatial self-similarity loss term to ensure the original content is unchanged. The generator $G$ and discriminator $D$ follow the architecture proposed in StyleGAN2 \cite{karras2020analyzing} and are used to compute the adversarial loss between $X^{\prime}_i$ and $X_j$. Texture co-occurrence loss between $X_i$ and $X_{j, b_i \neq b_j}$ is computed by the patch-discriminator $D_{patch}$, and the spatial self-similarity loss between $X_i$ and $X^{\prime}_i$ is computed by the $VGG$ model, respectively.

Finally, the original dataset $\mathcal{D}$ is combined with its augmented version $\mathcal{D}^{\prime}$ to train a classifier that is robust to inconsistent bias representations for the classification, and can avoid bias present in the training data. Additionally, $\mathcal{D}^{\prime}$ with $\mathcal{D}$ can be employed to train a segmentation model for adaptation on a target domain.

\subsection{Image Generation using Content and Texture}
Generating a realistic image using the content information in $X_i$ and the texture of $X_j$ requires the extraction of specific types of features. To this end, we use two different encoders $E_c$ and $E_t$ to extract a content-encoded tensor $c_i$ and a texture-encoded vector $t_j$, respectively. Here, $G$ generates a texture transferred image $X^{\prime}_i$ by taking $c_i$ and $t_j$ as a constant and style vector following \who{Karras}{karras2020analyzing}, and discriminator $D$ enforces a non-saturating adversarial loss \cite{goodfellow2014generative} for generative training. The adversarial loss is defined as:
\begin{equation}
	\begin{array}{c}
		L_{GAN}(E_c,E_t,G,D)=\mathbb{E}_{X_i,X_j \sim \mathcal{D}, b_i \neq b_j}[-\log(D(G(c_i,t_j)))].
		\label{eq1}
	\end{array}
\end{equation}
However, this setting often fails to retain content information since the discriminator is heavily enforced to generate an image that preserves entangled content features in the target domain. Thus, we add additional constraints to ensure generator $G$ retains content information in $c_i$ and uses texture information in $t_j$. These constraints take the form of two additional modules and loss terms, \ie a texture co-occurrence loss and spatial self-similarity loss.

\subsection{Texture Co-occurrence Loss}
To encourage the transfer of texture information $t_j$ from $X_j$ to $X^{\prime}_i$ \ie $G(c_i,t_j)$, we employ a texture co-occurrence loss with a patch-discriminator $D_{patch}$ \cite{park2020swapping} that measures the texture difference between $X^{\prime}_i$ and $X_j$. The texture co-occurrence loss and patch discriminator $D_{patch}$ were initially proposed for image manipulation \cite{park2020swapping} to disentangle texture information from structure. $D_{patch}$ encourages the joint feature statistics to appear perceptually similar \cite{julesz1962visual,julesz1981textons,gatys2015texture,park2020swapping}. This is achieved by cropping multiple random patch sizes between $1/8$ to $1/4$ of the full image size, and feeding them into $D_{patch}$. In particular, we average the features in $X_j$ patches, concatenate with the features of $X^{\prime}_i$, and feed them to the last layers of $D_{patch}$ to calculate the discriminator loss. Consequently, $G$ is enforced to satisfy the joint statistics of low-level features for consistent texture transfer. Formally,
\begin{equation}
	\begin{array}{c}
		L_{texture}(E_c,E_t,G,D_{patch})=\mathbb{E}_{X_i,X_j \sim \mathcal{D}, b_i \neq b_j}[-\log(D_{patch}(crop(X^{\prime}_i),crop(X_j)))].
		\label{eq2}
	\end{array}
\end{equation}

\begin{figure}[!t]
	\centering
	\includegraphics[width=0.5\columnwidth]{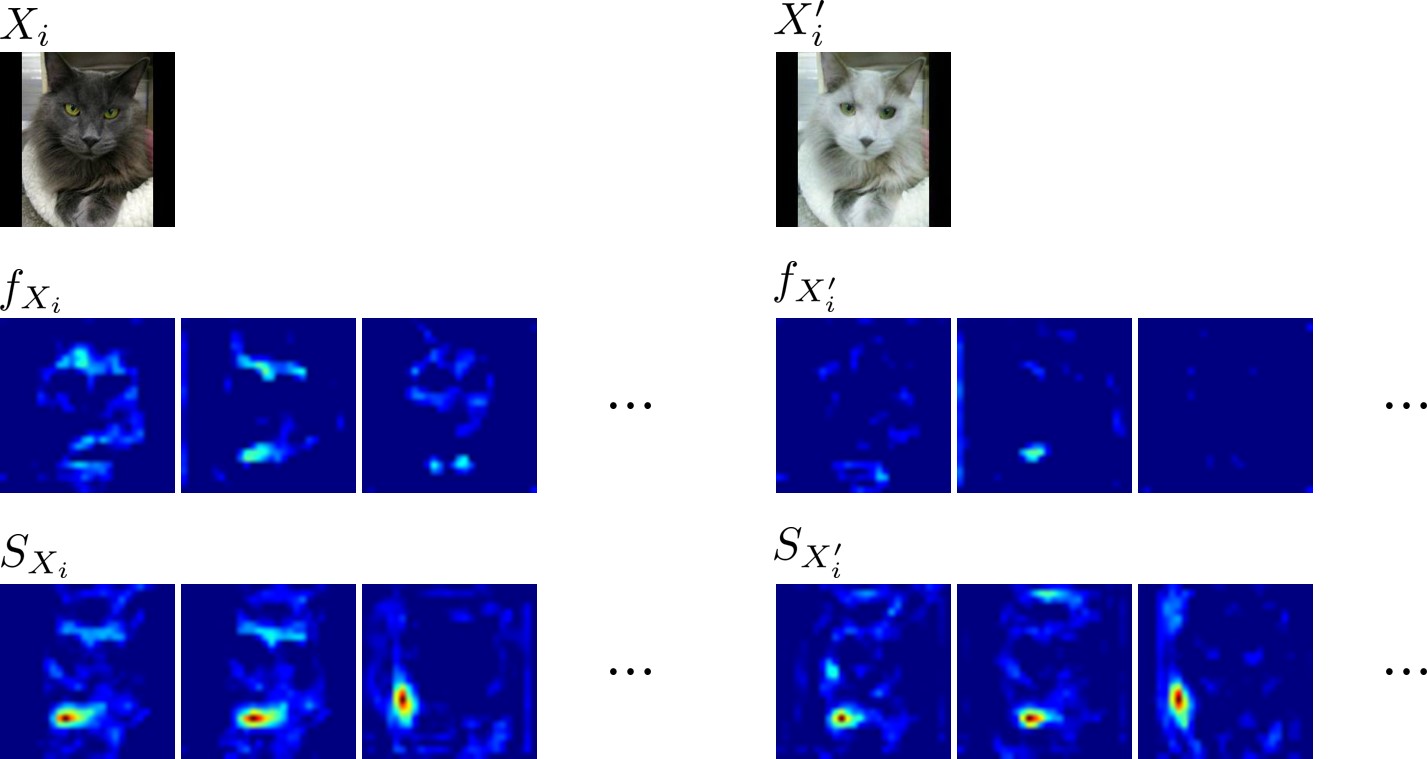}
	\caption{Visualization of intermediate feature maps obtained by $VGG$ and their self-similarity maps $S$. Even though $X_i$ and $X^{\prime}_i$ share the same contents, there are discrepancies between $f_{X_i}$ and $f_{X^{\prime}_i}$. On the other hand, self-similarity maps $S_{X_i}$ and $S_{X^{\prime}_i}$ are consistent.}
	\label{fig_feat}
\end{figure}

\subsection{Spatial Self-Similarity Loss}
To retain the content information of the source image, we employ spatial self-similarity as a domain invariant content constraint. Self-similarity loss has been used to maintain the structure of content images in artistic image style transfer~\cite{kolkin2019style}. Formally, a spatial self-similarity map is considered as follows:
\begin{equation}
	S_{X_i}=f_{X_i}^T \cdot f_{X_i},
\end{equation}
where $f_{X_i} \in \mathbb{R}^{C \times HW}$ denotes the spatially flattened features extracted from $VGG$ with channel $C$, height $H$, and width $W$, respectively. By applying the dot product, $S_{X_i} \in \mathbb{R}^{HW \times HW}$ captures spatial correlation from one location ($\mathbb{R}^C$) to the rest in all feature maps. Thus, our domain invariant content constraint (spatial self-similarity loss) can be calculated between $X^{\prime}_i$ and $X_i$ as follows:
\begin{equation}
	\begin{array}{c}
		L_{spatial}(E_c,E_t,G)=\mathbb{E}_{X_i,X_j \sim \mathcal{D}, b_i \neq b_j} [ \| 1 - cos(S_{X_i},S_{X^{\prime}_i}) \|_1 ],
	\end{array}
\end{equation}
where $cos$ denotes the cosine similarity. In conventional generative models, the reconstruction loss (\eg L1, MSE) or perceptual losses are used to provide constraints using the entangled feature (contents and texture) for minimizing content discrepancy (Figure \ref{fig_feat}). Since the content information can be decoupled from texture information using spatial self similarity, we can explicitly control the content features in $G$. Consequently, we can successfully preserve the content of $X_i$ while updating texture.

\subsection{Full Objective and Implementation Details}
The objective function of our framework is defined as:
\begin{equation}
	L_{total}=\lambda_{g} L_{GAN}+\lambda_{t} L_{texture}+\lambda_{s} L_{spatial},
\end{equation}
where hyper-parameters $\lambda_{g}$, $\lambda_{t}$, and $\lambda_{s}$ balance the importance of spatial and texture loss terms, respectively. We set $\lambda_{g} = 0.1$, $\lambda_{t} = 1.0$, and $\lambda_{s} = 100$, respectively. The objective and network design of discriminators $D$ and $D_{patch}$ closely follow StyleGAN2 \cite{karras2020analyzing}. $E_c$ and $E_t$ downsample their inputs 2$\times$ and 6$\times$ each to extract $c_i$ and $t_j$ (1$\times$ and 4$\times$ for small resolution inputs \eg digit), respectively. For $t_j$, we explicitly discard spatial information by applying global average pooling. To reduce computational costs, we select 256 random features from $f_{X_i}^T$ to obtain $\hat f_{X_i}^T$, thus reducing the size of the self-similarity map, \ie $\hat S_{X_i}= \hat f_{X_i}^T \cdot f_{X_i}$. Herein, $\hat S_{X_i} \in \mathbb{R}^{256 \times HW}$ is used to calculate $L_{spatial}$ instead of $S_{X_i} \in \mathbb{R}^{HW \times HW}$ (generally $256 < HW$). To construct $\mathcal{D}^{\prime}$, we randomly select texture sources from images with a different label.

\subsection{Extension to Multi-domain}
To debias a multi-domain biased dataset, the number of models required for texture updates is a factorial of the number of domains in the set. In other words, it is challenging to use the proposed method on datasets with a large number of labels (domains). Herein, we introduce a conditional version of the proposed method that can be constructed by using 2D embedding layers that simply change the statistics of an intermediate feature $\mathcal{F}$ of the CNN. The domain label is fed into a 2D embedding layer and returns 1D vectors (weight $e_w$ and bias $e_b$ in the same dimensions as $\mathcal{F}$) used for feature updates \ie $\mathcal{F}^\prime=\mathcal{F} \times e_w + e_b$. The updated feature $\mathcal{F}^\prime$ is fed to the next layer instead of $\mathcal{F}$. Consequently, a condition is provided to the model. We feed the 2D embedding layers to the second and pre-last CNN layers of $E_c$, $E_t$, and $D$. For $c_i$, $t_j$, and the cropped patches fed $G$ and $D_{patch}$, no condition is employed.

\section{Experiments}\label{sec:experiments}

In this section, we first illustrate the tasks and its corresponding datasets, and then describe the details of each experimental setting.
In Section \ref{sec:experiments_manipulation}, we describe the image manipulation task and its datasets to show the translation-characteristics of our method through comparisons against a state-of-the-art method.
In Section \ref{sec:experiments_debiasing}, we describe the debiasing task and its diverse datasets to demonstrate the superiority of image translation between domains with different semantics.
In Section \ref{sec:experiments_adaptation}, we describe the unsupervised domain adaptation task and its datasets to show the effectiveness and the advantageous of the structure preserving performance of the proposed method.

\subsection{Image Manipulation}\label{sec:experiments_manipulation}
\subsubsection{Task}
In this task, designers often modify images using photo-editing tools \eg Photoshop, which is often time-consuming and labor intensive. Thus, image manipulation using unsupervised machine learning is a topic of interest with practical uses. Existing methods commonly disentangle content (structure) and texture, and then combine these features to generate a new image. The later can be similarly achieved via our proposed method \ie employ self-similarity to preserve the source image's content without the interference of texture in the source image. In our evaluation of this task, we compare against a state-of-the-art image manipulation method with ablation studies to highlight visual characteristic differences, and provide more evidence on the advantages of our method.

\subsubsection{Datasets}
\textbf{LSUN Churches} \cite{yu2015lsun} and \textbf{Animal Faces HQ (AFHQ)} \cite{choi2020stargan} have been used to verify image manipulation performance. We crop images to 256$\times$256 during training; during testing, we resize the short side of the image to 256 while maintaining the aspect ratio following \cite{park2020swapping}.

\subsubsection{Implementation Details}
We compare our method against a state-of-the-art image-manipulation method \cite{park2020swapping}, and include ablation studies to assess the impact of the proposed components by omitting each loss function. For fair evaluation, we follow the same settings described in \cite{park2020swapping} except the batch size (we set 16).

\subsection{Debiasing Classification}\label{sec:experiments_debiasing}
\subsubsection{Task}
Classification models trained on datasets with texture bias usually perform poorly on out-of-distribution samples since biased representations are embedded into the model. Solving the bias problem is a vital requirement for most medical and industrial domains that have strict structural constraints on their images, and cannot solely rely on web or social media to collect a large number of data samples. We can leverage our proposed method to obtain the debiased classifier by adjusting the distribution of the training dataset. Specifically, our method translates images while maintaining the structure of the source input, even if the semantics of the source and target images are different. Hence, our method is expected to adjust the training dataset distribution more efficiently for debiasing.

\begin{figure}[!t]
	\centering
	\includegraphics[width=1.0\columnwidth]{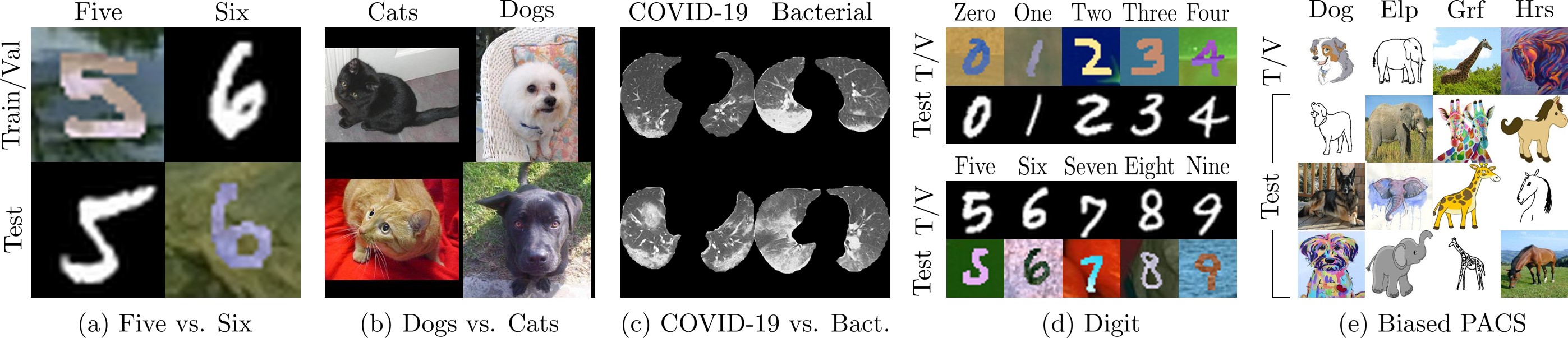}
	\caption{Examples of texture biased datasets. T/V indicates Train/Val. Elp, Grf, and Hrs in Biased PACS (e) indicates elephant, giraffe, and horse.}
	\label{fig_dataset}
\end{figure}

However, existing datasets often have similar distributions in both training and testing data \cite{barbu2019objectnet,bissoto2020debiasing}, making them unsuitable for direct bias mitigation analysis. Thus, we constructed five texture biased datasets following previous works \cite{alvi2018turning,kim2019learning}, where the training and testing sets have opposite distributions allowing us to solely focus our evaluation on texture bias mitigation (Fig. \ref{fig_dataset}). In particular, COVID-19 vs. Bacterial pneumonia is a dataset that can verify whether the proposed method can solve real-world medical problems.

\subsubsection{Datasets}
\textbf{Five vs. Six} is based on MNIST \cite{lecun2010mnist} and MNIST-M \cite{ganin2015unsupervised} datasets where only the numbers five and six are used to construct a training split.
Images with a five are taken solely from MNIST-M, whereas those with six are from MNIST, the opposite case is considered for the testing split.

\textbf{Dogs vs. Cats} was originally proposed by \who{Kim}{kim2019learning} by aligning the hair color differences of animals.
Images with bright dogs and dark cats images are used in the training split, whereas dark dogs and bright cats are employed for testing. Consequently, the dataset exhibits hair color bias.

\textbf{COVID-19 vs. Bacterial pneumonia} is constructed to evaluate our method on real-world medical problems.
Pneumonia caused by different pathogens requires specific treatments. However, images obtained by computed tomography (CT) scans might share similar properties making accurate diagnosis challenging. One source of ambiguity for automatic diagnostic systems is the choice of the CT protocol during the image acquisition process, \ie using a contrast agent vs. another protocol. To evaluate our method for real-world texture bias, CT scans were collected at Yeungnam University Medical Center. In this dataset, we selected Non-contrast COVID-19 for training and contrasted scans for testing. The opposite applies for Bacterial pneumonia scans.

To evaluate our method on \textbf{Multi-class Biased Datasets}, we constructed \textbf{Digit} and \textbf{Biasd PACS} datasets.
In the Digit dataset, samples with labels from zero to four are taken from MNIST \cite{lecun2010mnist}, and those with five to nine from MNIST-M \cite{ganin2015unsupervised}. Biased PACS dataset was constructed using PACS dataset \cite{li2017deeper}, which consists of four domains (Photo, Art, Cartoon, Sketch) and has seven classes. We selected the top four labels (Dog, Elephant, Giraffe, Horse) after sorting the number of images. Each class is taken from a different domain (\eg Dog - Cartoon, Elephant - Sketch) for the training dataset, and the remaining domain images are used for the test set. Additionally, we constructed an \textbf{Inverse Biased PACS} dataset that inversely uses the biased PACS dataset, \ie replace the test split samples with the training samples, and replace the train/val split samples with the testing samples, respectively. Compared to the binary-class biased datasets, this dataset has access to samples in three domains during training and be evaluated on samples from a single domain that does not match the training data.

\subsubsection{Implementation Details}
We compare our method against a baseline \cite{he2016deep}, five conventional debiasing methods \cite{louppe2017learning,alvi2018turning,zhang2018mitigating,wang2019balanced,kim2019learning}, two domain generalization methods \cite{nuriel2021permuted,zhou2021domain}, and five image translation models \cite{zhu2017unpaired,liu2017unit,huang2018munit,lee2020drit++,park2020cut}. Image size was $224 \times 224$ for Dogs vs. Cats, COVID-19 vs. Bacterial pneumonia, Biased PACS, and Inverse Biased PACS datasets, and $32 \times 32$ for Five vs. Six and Digit datasets. For our \textbf{baseline}, we train an ImageNet pre-trained ResNet50\cite{he2016deep} classifier only using the biased training data without any debiasing methods.
For non-generative \textbf{debiasing} methods, bias information (\eg data-source, color, and CT protocol) was used to train the classifier. For fair evaluation, the classifier in these methods were trained using the same backbone (ResNet50) under the same training settings. For \textbf{image translation} methods, we constructed augmented datasets using each model and employed them to train classifiers.
In the classification step, the settings for classifier were the same as the baseline.
For \textbf{domain generalization} models, the best hyper-parameters reported in their paper were used for training (\eg alpha in \cite{zhou2021domain} and probability in \cite{nuriel2021permuted}), other settings were left the same as the baseline classifier.
We evaluated accuracy using the macro F1-score, which treats the label distribution equally. For COVID-19 vs. Bacterial pneumonia, we aggregated slice predictions via majority voting to obtain patient-level diagnosis. For consistency, training was repeated three times, and we report the average performance with the standard deviation of each method as the final performance. In the multi-domain data scenario, we did not perform experiments for methods \cite{louppe2017learning,zhang2018mitigating} addressing binary classification, as well as generation methods \cite{zhu2017unpaired,liu2017unit,huang2018munit,lee2020drit++,park2020cut} that had to learn models for all domain labels.

\subsection{Unsupervised Domain Adaptation}\label{sec:experiments_adaptation}
\subsubsection{Task}
Constructing semantic segmentation labels is extensively time-consuming and laborious. Therefore, employing synthetic images from graphics game engines for dataset construction is common approach, but due to image discrepancy between synthetic and real data, model performance degrades significantly \cite{hoffman2018cycada,tsai2018learning}. To alleviate these issues, various unsupervised domain adaptation methods have been proposed. These methods try to satisfy the hypothesis that the higher the performance will be obtained when the better the image(or feature) is translated into the target domain. The proposed method can generate an image similar to the target domain by using texture information from the target. In addition, compared to the method that uses a trained segmentation model for content maintenance \cite{li2019bidirectional}, our method is more efficient since it does not require a segmentation model to preserve content. Furthermore, the previous unsupervised domain adaptation method relied heavily on the DeepLabV2 \cite{chen2017deeplab}, and burden of works are required to integrate adaptation modules to the new segmentation model and needed to find a hyper-parameter that works well. However, since we can train a target adapted model through replacing a training dataset, our method can be applied recent state-of-the-art segmentation models in a more tractable way.

Additionally, due to performance drops arising from the image acquisition process \eg the difference in machine and scanning protocols, we believe it is straightforward to employ image translation methods for this task. Specifically, alignment based unsupervised domain adaptation methods are vital to be used in image translation methods, comparison against translation methods indicate verifying the image-level alignment performance solely. Since the acquisition difference causes the positional and scale mismatches of the interest (\eg prostate) as well as texture differences, research interest in content preservation is increasing. Note that if content is excessively deformed after translation, the produced segmentation mask will equally be poor, leading to low segmentation performance. Thus, to fairly compare with translation models, we used images $\mathcal{X}^A$ and $\mathcal{X}^B$ from domains $A$ and $B$. In particular, we translated $\mathcal{X}^A$ to $\mathcal{X}^{A \rightarrow B}$ so that the texture of $\mathcal{X}^{A \rightarrow B}$ is similar to $\mathcal{X}^B$. The segmentation model is then trained using both $\mathcal{X}^A$ and $\mathcal{X}^{A \rightarrow B}$, and then evaluated on $\mathcal{X}^B$.

\subsubsection{Datasets}
\textbf{GTAV-Cityscapes} is employed to evaluate our method on a real-world unsupervised domain adaptation task.
Cityscapes \cite{cordts2016cityscapes} dataset was proposed for semantic urban scene understanding, and GTAV \cite{richter2016playing} is a synthetic dataset that has overlapping/compatible semantic categories with Cityscapes that rendered by the gaming engine Grand Theft Auto V.
We use all the official 19 training classes in our experiments.

\textbf{Prostate} is constructed to evaluate our method on a medical unsupervised domain adaptation task. We employed prostate T2-weighted MRI images from two different domains (A: BIDMC, B: I2CVB) \cite{bloch2015nci}.
Segmentation masks were provided for prostate regions. In our pre-processing pipeline, the prostate regions (in axial plane) were center-cropped and resized to 256$\times$256.

\subsubsection{Implementation Details}
For \textbf{GTAV-Cityscapes}, we compare our method against state-of-the-art unsupervised domain adaptation methods.
For image translation, we followed the same training settings \cite{hoffman2018cycada}.
For the segmentation model, we employed a state-of-the-art segmentation model \ie SegFormer \cite{xie2021segformer} with a mmsegmentation \cite{mmseg2020} framework.
For evaluation, We follow the same protocol \cite{hoffman2018cycada,tsai2018learning}, \ie train the model on the unlabeled training set and report the results on the validation set.

For \textbf{Prostate} experiments, we compare our method against five image translation models \cite{zhu2017unpaired,liu2017unit,huang2018munit,lee2020drit++,park2020cut}.
For each method, we first apply the corresponding model to generate additional images (\eg $\mathcal{X}^{A \rightarrow B}$), train a segmentation model (U-Net \cite{ronneberger2015u}) using source and generated images (\eg $\mathcal{X}^{A}$ and $\mathcal{X}^{A \rightarrow B}$), and finally tested the segmentation model on the target domain images (\eg $\mathcal{X}^{B}$). Since the dataset contains two domains, the same procedure is repeated in the opposite way and the average performance was compared. The segmentation performance was evaluated by the Dice similarity coefficient (DSC) score. We trained the segmentation models for three times with deterministic training using different seeds (=0,1,2), and reported the final score by averaging the test DSC score on the models with the best validation scores.



\begin{figure}[!t]
\begin{minipage}[t]{0.5\linewidth}
	\centering
	\includegraphics[width=1.0\columnwidth]{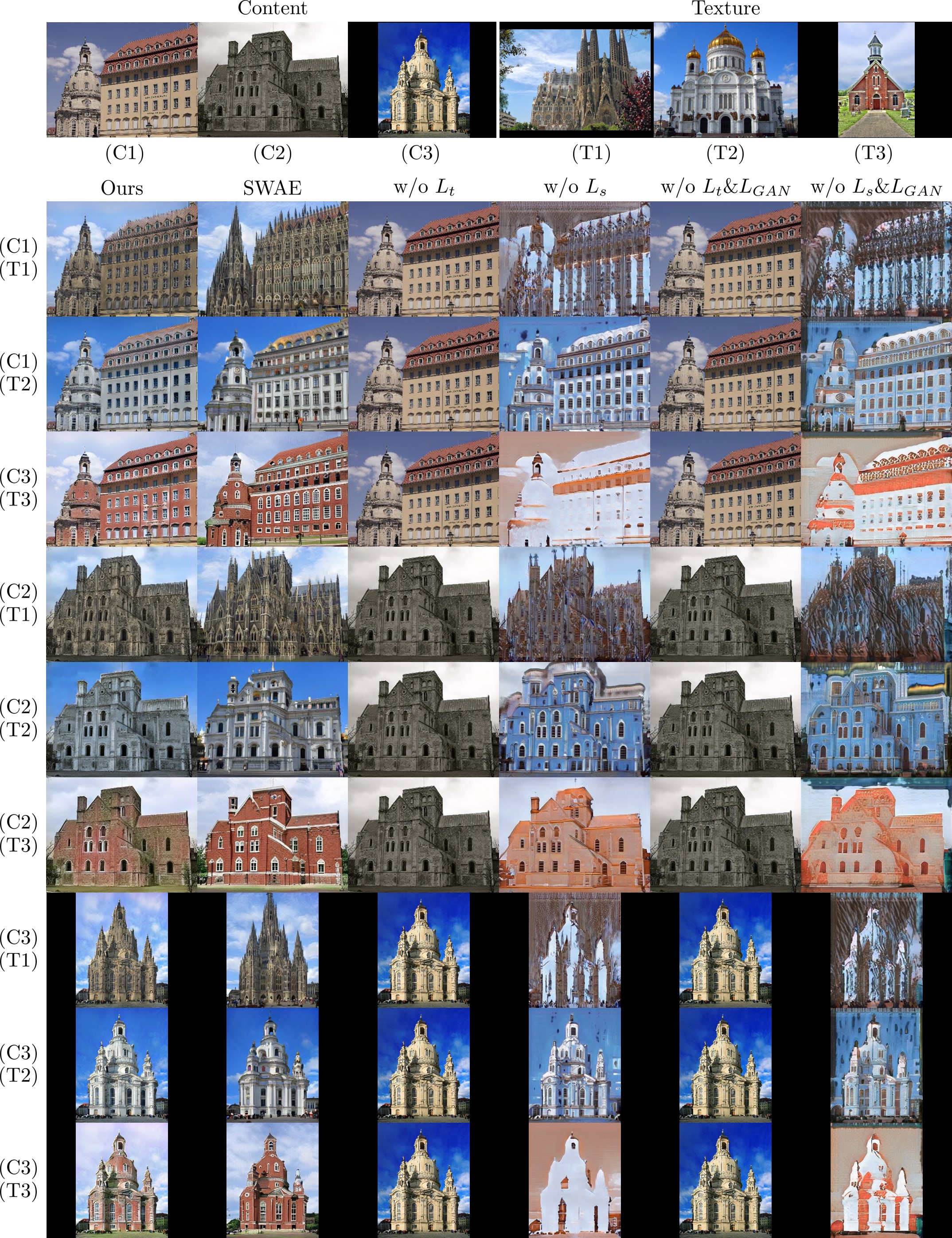}
    \caption{Qualitative results of the proposed method, the swapping-autoencoder (SWAE), and ablation studies on the LSUN Churches dataset. $L_s$ and $L_t$ indicates $L_{spatial}$ and $L_{texture}$. The source and target images for this experiment are borrowed from the figures of the competitor's paper \cite{park2020swapping}.}
	\label{fig_manipulation_church}
\end{minipage}
\hspace{0.1cm}
\begin{minipage}[t]{0.5\linewidth}
	\centering
	\includegraphics[width=1.0\columnwidth]{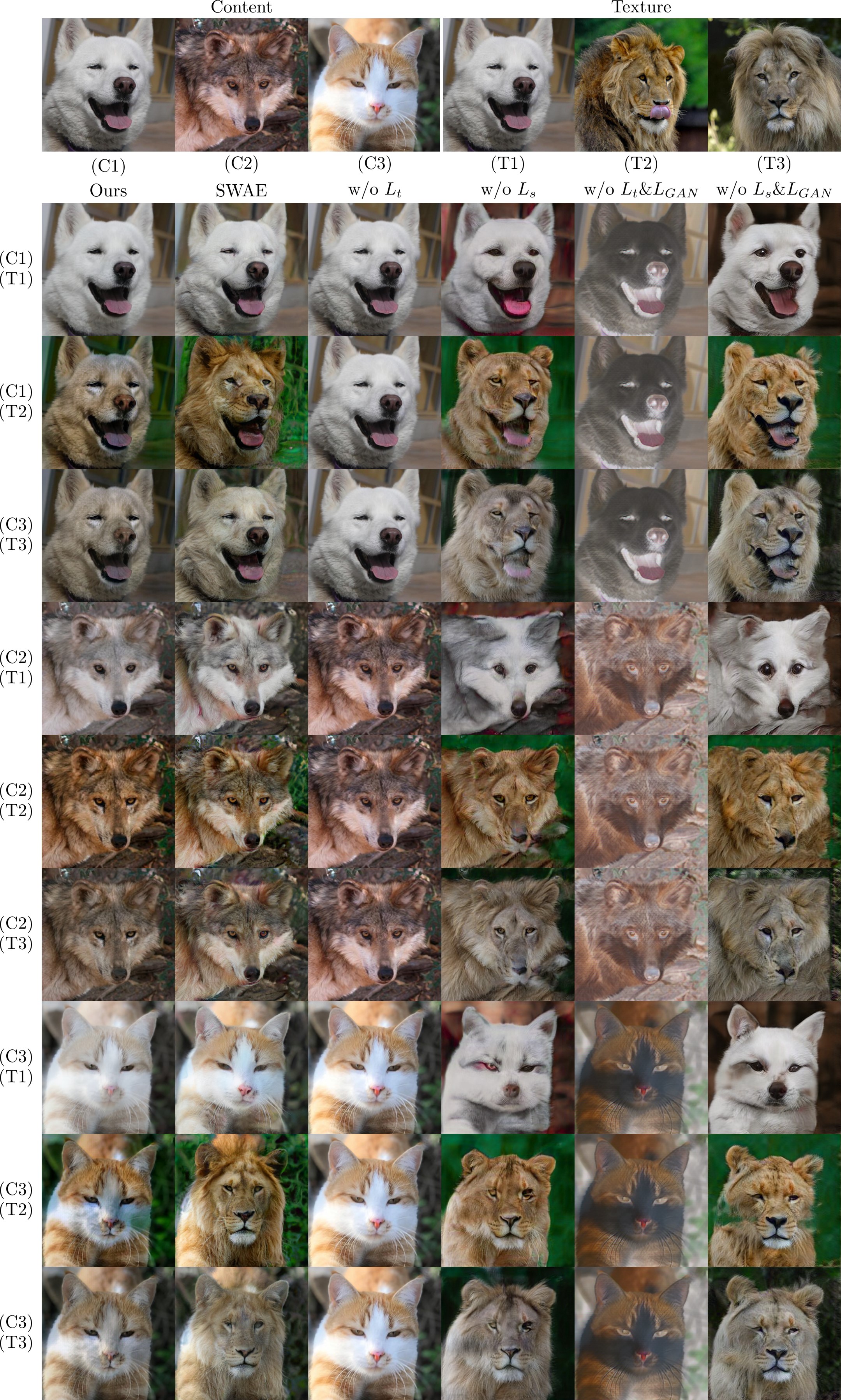}
    \caption{Qualitative results of the proposed method, the swapping-autoencoder (SWAE), and ablation studies on the AFHQ dataset. $L_s$ and $L_t$ indicates $L_{spatial}$ and $L_{texture}$. The source and target images for this experiment are borrowed from the figures of the competitor's paper \cite{park2020swapping}.}
	\label{fig_manipulation_afhq}
\end{minipage}
\end{figure}

\section{Results}\label{sec:results}

\subsection{Image Manipulation}s
For this task, our method shows improved content preservation on the Churches dataset (see Fig. \ref{fig_manipulation_church}) compared to swapping-autoencoder (SWAE) \cite{park2020swapping}; a state-of-the-art image manipulation method.
For instance, extensive content distortions were observed on the source images when SWAE transferred steeple texture to the roof of the source image (see. Fig. \ref{fig_manipulation_church} with (T1)). In addition, Fig. \ref{fig_manipulation_church} with (T2) shows window distortions, with new windows created on the roof and the chimney, as well as Fig. \ref{fig_manipulation_church} with (T3) that shows window distortions, and has a missing little steeple. In contrast, the proposed method maintains most of the source image's content i.e., details of the roof, window, and steeple are retained with stable texture transfer.

The results on AFHQ dataset are shown in Fig. \ref{fig_manipulation_afhq}, both methods achieved reasonable image translation without content distortion. However, SWAE (Fig. \ref{fig_manipulation_afhq}(C1)(T2),(C3)(T2),(C3)(T3)) generated images with the structure of the texture images \textit{i.e.,} generating an image of a lion instead of the dog or cat. In contrast, our method does not change the species of the content image while updating texture, and is considered to be the main reason for the high debiasing and adaptation performance reported in Tables \ref{tab_classification}, \ref{tab_segmentation_gtav_cityscapes}, and \ref{tab_segmentation_prostate}, respectively.

We also conducted ablation studies by separating each module to establish which characteristics contribute the most to our target goal.
Ablation results on Churches dataset are shown in Fig. \ref{fig_manipulation_church}.
The absence of $L_{texture}$ and $L_{texture}$\&$L_{GAN}$ failed to update texture as similar images were generated.
Moreover, content information is significantly distorted when $L_{spatial}$ and $L_{spatial}$\&$L_{GAN}$ when excluded.
For instance, the methods applied the texture of the sky on a building in reverse, showing that the model can not distinguish the semantics in each image.
Consequently, this shows that $L_{spatial}$ is vital for realistic image generation and not only for preserving spatial information when content between images significantly differs.

In addition, similar results were observed on AFHQ dataset (Fig. \ref{fig_manipulation_afhq}).
The model without $L_{texture}$ generates an identical image without texture updates.
When excluding $L_{texture}$ \& $L_{GAN}$, though the content information is preserved correctly, the color of generated image is not realistic.
When $L_{spatial}$ is absent, shapes were transformed into those of texture-source-like species (\eg more rounded ears), and additional distortions are observed when $L_{GAN}$ was excluded as the eye became larger than that of content source species.
In summary, our ablation studies demonstrate that each proposed module contributes to texture transfer and content preservation as intended.

\begin{table}[!t]
    \centering
    \small
    \caption{Classification performance of non-generative debiasing methods (The first sub-row), domain generalization methods (The second sub-row), image translation models (The third sub-row), and the proposed model (The fourth sub-row) on five datasets. F1-score with $\pm$ std was used as the metric. The second and third sub-columns indicate binary- and multi-class biased dataset classification performances, respectively.}
    \label{tab_classification}
    \begin{tabular}{l||c|c|c||c|c|c}
        \hline
        Method & (a) Five vs. Six & (b) Dogs vs. Cats & (c) COV. vs. Bac. & (d) Digit & (e) B. PACS & ($e^\prime$) In. B. P. \\
        \hline \hline
        Baseline & ${1.81 \pm 0.04}$ & ${67.27 \pm 2.21}$ & ${48.63 \pm 9.81}$ & ${3.96 \pm 1.47}$ & ${12.89 \pm 2.84}$ & ${29.01 \pm 2.04}$ \\  
        \hline
        Learn to Pv. & ${1.42 \pm 0.52}$ & ${73.85 \pm 2.20}$ & ${66.53 \pm 7.59}$ & - & - & - \\
        \hline
        Adv. Debias & ${1.27 \pm 0.21}$ & ${72.42 \pm 3.01}$ & ${59.71 \pm 6.57}$ & - & - & - \\ 
        \hline
        BlindEye & ${1.60 \pm 0.76}$ & ${77.13 \pm 4.72}$ & ${65.75 \pm 4.07}$ & ${3.93 \pm 1.26}$ & ${10.72 \pm 1.34}$ & ${28.33 \pm 3.24}$ \\ 
        \hline
        Not Enough  & ${1.82 \pm 1.02}$ & ${72.06 \pm 3.49}$ & ${61.57 \pm 5.24}$ & ${6.20 \pm 3.23}$ & ${10.23 \pm 1.31}$ & ${37.58 \pm 12.45}$ \\ 
        \hline
        LNTL  & ${2.85 \pm 1.91}$ & ${69.35 \pm 5.78}$ & ${61.57 \pm 5.24}$ & ${11.26 \pm 1.91}$ & ${12.16 \pm 2.46}$ & ${34.04 \pm 5.76}$ \\
        \hline \hline
        P-AdaIN & ${1.67 \pm 0.70}$ & ${73.21 \pm 4.60}$ & ${59.71 \pm 6.57}$ & ${1.85  \pm 0.61}$ & ${11.83 \pm 1.06}$ & ${37.46 \pm 5.89}$ \\
        \hline
        MixStyle & ${16.47 \pm 9.36}$ & ${72.51 \pm 3.08}$ & ${55.78 \pm 7.96}$ & ${5.54 \pm 3.28}$ & ${10.39 \pm 0.96}$ & ${34.80 \pm 1.44}$ \\
        \hline \hline
        CycleGAN & ${23.07 \pm 2.47}$ & ${89.66 \pm 1.20}$ & ${59.33 \pm 2.42}$ & ${1.90 \pm 0.21}$ & - & - \\ 
        \hline
        UNIT & ${1.09 \pm 0.65}$ & ${85.80 \pm 3.10}$ & ${59.96 \pm 10.04}$ & ${6.30 \pm 0.84}$ & - & - \\ 
        \hline
        MUNIT & ${1.48 \pm 0.33}$ & ${86.81 \pm 1.81}$ & ${65.54 \pm 7.93}$ & ${8.61 \pm 1.05}$ & - & - \\ 
        \hline
        DRIT++ & ${31.85 \pm 2.34}$ & ${80.87 \pm 1.61}$ & ${47.07 \pm 9.91}$ & ${30.80 \pm 0.93}$ & - & - \\ 
        \hline
        CUT & ${6.50 \pm 0.78}$ & ${86.19 \pm 1.22}$ & ${52.75 \pm 15.45}$ & ${2.96 \pm 0.72}$ & - & - \\
        \hline \hline
        Proposed & \boldmath{${75.18 \pm 3.52}$} & \boldmath{${91.80 \pm 0.44}$} & \boldmath{${76.50 \pm 3.38}$} & \boldmath{${68.82 \pm 1.27}$} & - & - \\
        \hline
        Proposed (Ext.) & ${72.62 \pm 2.56}$ & ${90.80 \pm 0.59}$ & ${70.70 \pm 5.80}$ & ${63.82 \pm 0.71}$ & \boldmath{${36.68 \pm 1.24}$} & \boldmath{${46.48 \pm 4.09}$} \\
        \hline
    \end{tabular}
\end{table}

\begin{figure}[!t]
\centering
	\includegraphics[width=0.5\columnwidth]{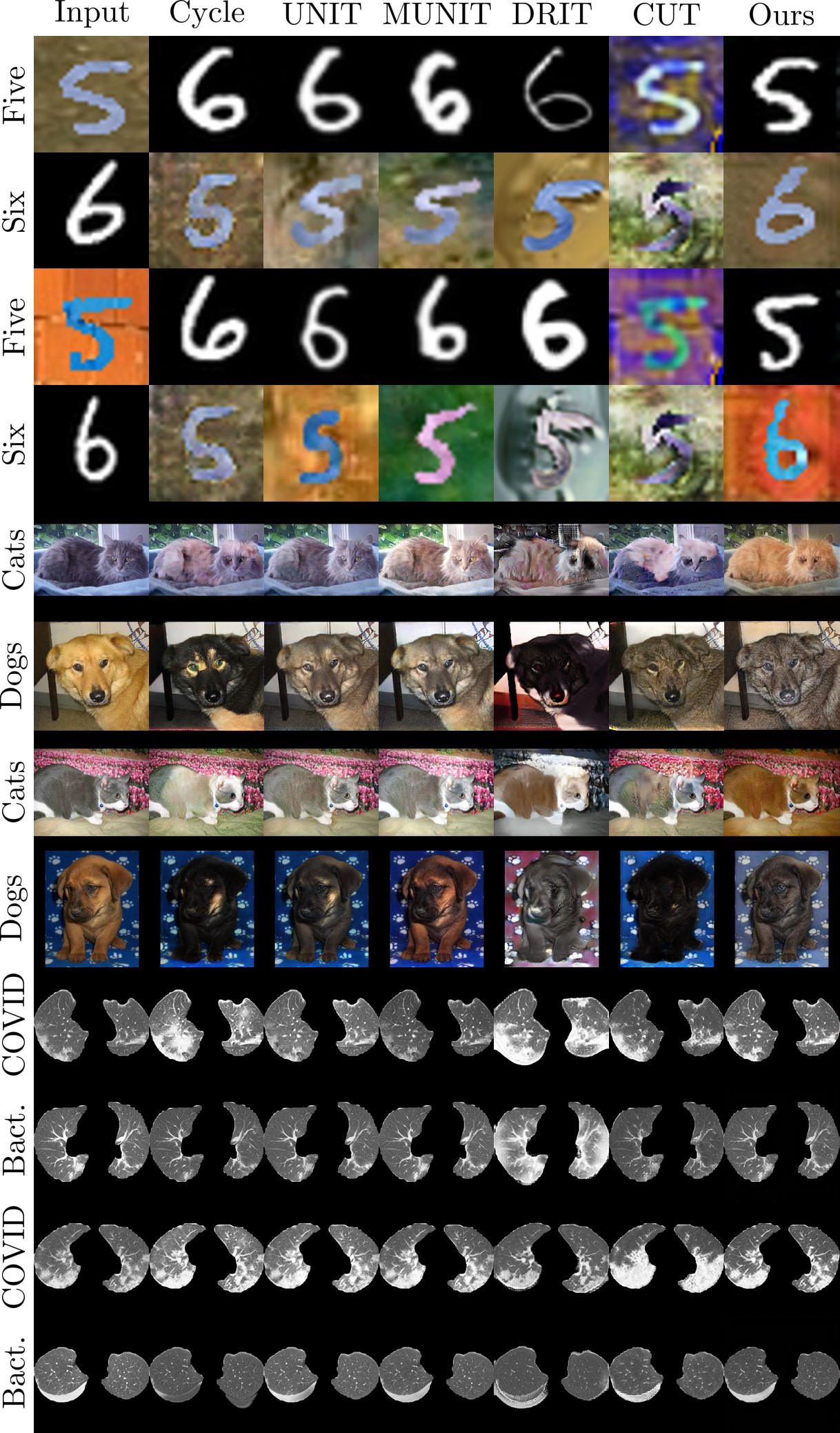}
	\caption{Qualitative results of five image translation models and the proposed on Five vs. Six, Dogs vs. Cats, and COVID-19 vs. Bacterial Pneumonia dateasets. Our model uses the content from the source image, and the texture from an image with a different label. For each dataset, same row image is used as source, whereas the texture is from the opposite pair. Our model successfully transfers texture while retaining the content.}
	\label{fig_classification_qualitative_large}
\end{figure}

\subsection{Debiasing Classification}
\subsubsection{Five vs. Six}
In this task, none of the compared models yielded satisfactory results (see Table \ref{tab_classification}(a)). Non-generative models report extremely low scores, leading us to conclude that texture features were still used for classification instead of the digit shape. Furthermore, domain generalization methods also show poor F1-scores. We observed that image translation models mainly transferred texture and shape jointly (See Fig. \ref{fig_classification_qualitative_large}). Consequently, the resulting augmented dataset had several instances with the shape of a different category. As the classifier does not have a clear cue to distinguish between numbers, this led to poor performance. Results on this dataset clearly show the benefit of using our method to transfer texture features while retaining the underlying information for successful bias mitigation.

\subsubsection{Dogs vs. Cats}
Performing classification on real-world animals requires more complex features than color alone. In Table \ref{tab_classification}(b), we observed an increase in performance when a debiasing- or domain generalization method was applied. Overall, image translation models show better performance over the non-generative counterparts, \ie F1-score +10\%. While the performance of image translation models was reasonable, several limitations were noted. In the case of CycleGAN, CUT, and DRIT++, unsatisfactory images were obtained, \ie texture was not entirely translated and distortions were present. Moreover, UNIT and MUNIT generate images with small texture updates, thus bias remains in the generated images (see Fig. \ref{fig_classification_qualitative_large}). On the other hand, our model simultaneously shows improved performance with high-quality image generation. We believe this is mainly due to the proposed texture translation strategy. Leveraging both texture co-occurrence and spatial self-similarity losses enabled our model to generate consistent and natural images leading to improved classification performance.

\subsubsection{COVID-19 vs. Bacterial Pneumonia} In contrast to the results observed on the Dogs vs. Cats task, non-generative models report improved performance over the rest (Table \ref{tab_classification}(c)). Domain generalization methods obtain lower F1-scores than non-generative approaches, whereas non-generative models mitigate distinctly recognizable bias in the image such as color over image translation models, since they do not impose any structural changes in the image. On the other hand, image translation models tend to modify patterns on regions of diagnostic interest, thus losing the properties that identify the disease, leading to lower performance (See Fig. \ref{fig_classification_qualitative_large}). For example, CycleGAN tends to erase (row 10, 12) or create lesions (row 9) in the CT scans. UNIT and MUNIT show minor texture updates, hence are insufficient to mitigate bias in the classifier during training (all rows). DRIT++ and CUT create artifacts such as non-existent lesions, checkerboards and change the images' properties (row 9, 10, 12), resulting in the lowest classification accuracy among the image translation models. Meanwhile, our method jointly optimizes the texture co-occurrence and spatial self-similarity losses, each imposing structural and texture constraints for image generation. Herein, our method can successfully update texture without introducing artifacts in the original CT scan. Note that while contrast CT is a standard protocol for common lung disease diagnosis, curating contrast CT is challenging since extra processes such as contrast agent injection and disinfection are required \cite{pontone2020role,kalra2020chest}, especially during the pandemic. Finally, we believe texture biases will be unexpectedly or sometimes unavoidably be introduced during the data collection process; thus, a debiasing method that can maintain key structures in their images is vital in the medical domain.

\begin{figure}[!t]
	\centering 
	\includegraphics[width=0.5\columnwidth]{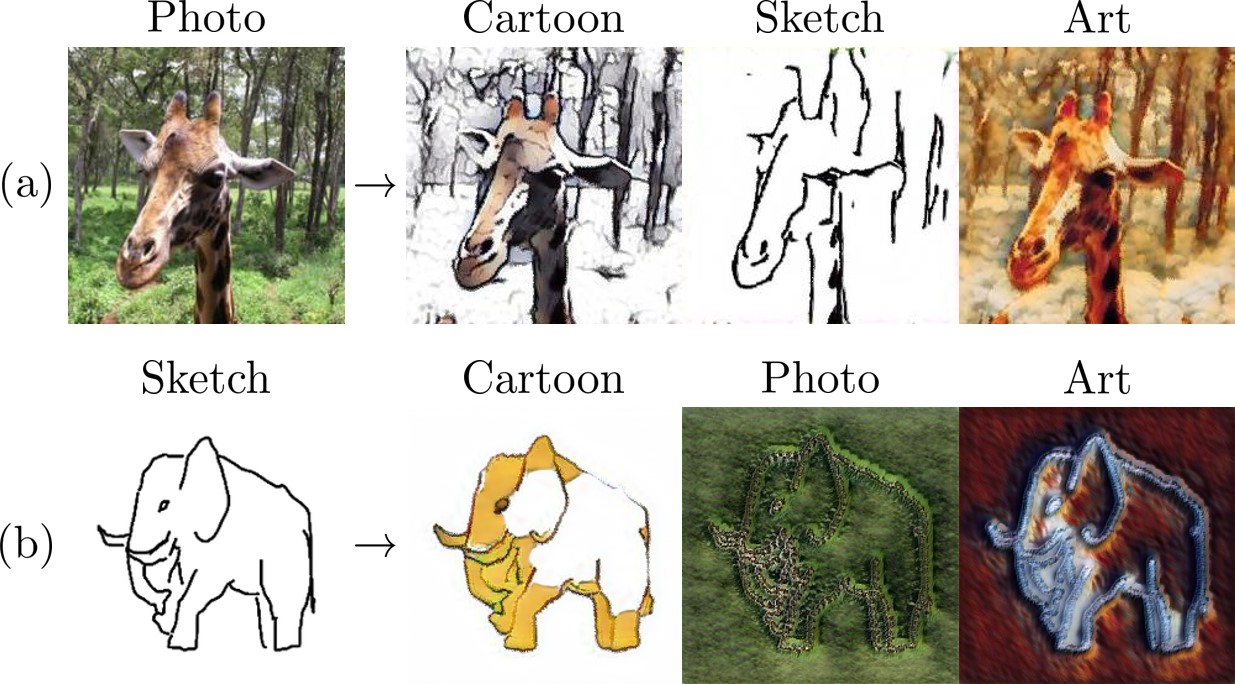}
	\caption{Qualitative results of the proposed method on the Biased PACS dataset. (a) Cartoon, sketch, and art images are generated from a photo image. (b) Cartoon, photo, and art images are generated from a sketch image.}
	\label{fig_classification_qualitative_pacs}
\end{figure}

\subsubsection{Multi-class Biased Datasets}
For the multi-class Digit dataset, none of non-generative models and domain generalization methods report satisfactory performance (see Table \ref{tab_classification}(d)). On the other hand, our method achieved the best score over all compared methods (\ie 68.82\%). On Biased PACS dataset, all comparison methods report F1-scores around 10\%, while our method significantly improved to 36.68\% (Table \ref{tab_classification}(e)). The translated images from Photo to other domains are satisfactory as shown in Fig. \ref{fig_classification_qualitative_pacs}(a), whereas Sketch to Photo (and Art) have differences between the realistic Photo (and Art) images as shown in Fig. \ref{fig_classification_qualitative_pacs}(b). We found that this is due to the scale of Biased PACS dataset (small with only a few hundred images). Despite this, our method achieves reasonable generation quality, retains content and accurately transfers textures for debiasing. On Inverse Biased PACS dataset, even though non-generative models and domain generalization methods utilize multiple domain samples for training, they perform poorly on out-of-distribution samples as shown Table \ref{tab_classification}($e^\prime$). This highlights the difficulty of learning domain agnostic features. In summary, we empirically show that images generated with our method can help the classifier mitigate texture biases towards improved performance, and can be applied to not only binary-domain \& binary-class classification, but also multi-domain \& multi-class classification tasks.

\subsubsection{Ablation Study} We performed ablation studies to evaluate the impact of the content $L_{spatial}$ and texture $L_{texture}$ losses by either removing one or replacing them with a different loss function. It is essential to validate whether texture co-occurrence and self-similarity are the key components for improving image generation quality in biased settings. Thus, we used style \cite{li2017demystifying} and perceptual \cite{gatys2016image} losses to replace $L_{spatial}$ and $L_{texture}$, respectively, as they are the most common techniques used in style transfer methods to retain content features for image generation. On the other hand, the adversarial $L_{GAN}$ loss is an essential part of our image generation pipeline and unrealistic results were produced when omitted, thus we include this loss in our ablations.

\begin{figure}[!t]
	\centering
	\includegraphics[width=0.5\columnwidth]{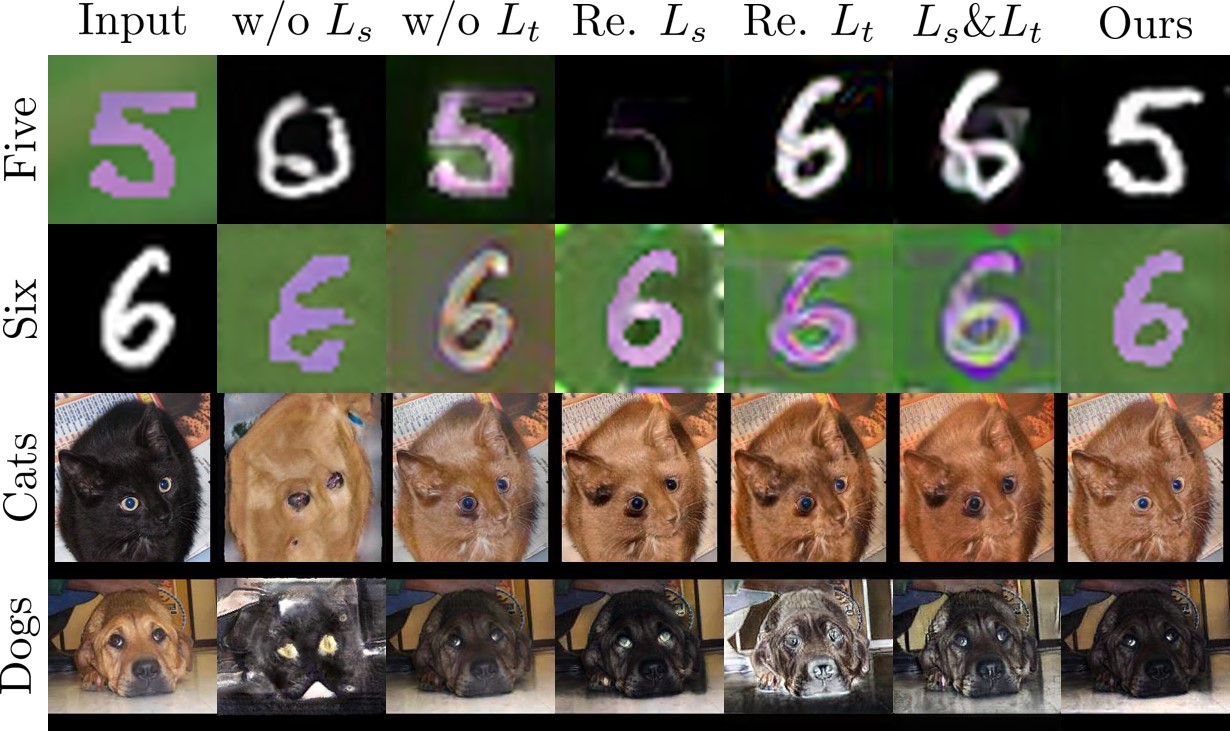}
	\caption{Qualitative results on ablation studies. $L_s$ and $L_t$ indicates $L_{spatial}$ and $L_{texture}$. Re. $L_s$, Re. $L_t$, and $L_s\&L_t$ indicate replacing with perceptual loss, style loss, and perceptual \& style loss for each loss function.}
	\label{fig_classification_ablation}
\end{figure}

Results on the Five vs. Six dataset in Table \ref{tab_classification_ablation}(a) show that the use of texture co-occurrence is a key in enabling the classifier to correctly differentiate digits. However, qualitative results in Fig. \ref{fig_classification_ablation} indicate that the absence of $L_{texture}$ results in a failure to change the texture and correctly mitigate bias information. Likewise, replacing $L_{texture}$ with a style loss overcomes the content constraints by modifying both texture and shape. In addition, while the absence of $L_{spatial}$ lets the classifier achieve higher performance than most methods, qualitative results show that even though the texture is correct, the shape does not resemble neither a number five nor six, which can be exploited by the binary classifier and can lead to incorrect predictions. Moreover, replacing $L_{spatial}$ or $L_{spatial}$ \& $L_{texture}$ results in poor image quality with ambiguous shapes. Consequently, we show that the combination of texture co-occurrence and self-similarity losses leads to higher image quality towards mitigating the underlying texture bias.

For Dogs vs. Cats, the model seems to be less susceptible to a drop in performance due to a change in the loss term, and usually generates good quality images as is reflected by the good classification results. The only exception is the absence of $L_{spatial}$, which reported a lower score (Table \ref{tab_classification_ablation}(b)), and it is clear from Fig. \ref{fig_classification_ablation} that the shape is lost in a similar fashion to the results observed in Five vs. Six data. In contrast, COVID-19 vs. Bacterial pneumonia results show high dependency on the loss functions leading to decrease in performance across all ablation studies. We believe this due to the explicit constraints of content and texture, as they are important to reduce the risk of unintended distortion, and insufficient texture translation, respectively.

\begin{table}[!t]
    \centering
    \small
    \caption{Classification performance of the ablation study of the proposed model using F1-score with $\pm$ std on three datasets. w/o $L_s$ and w/o $L_t$ indicate training without $L_{spatial}$ and $L_{texture}$, respectively. Replace $L_s$, Replace $L_t$, and Replace $L_s$ \& $L_t$ indicate replacing with perceptual loss \cite{gatys2016image}, style loss \cite{li2017demystifying}, and perceptual \& style loss for each loss function.}
    \label{tab_classification_ablation}
    \begin{tabular}{l||c|c|c}
        \hline
        Method & (a) Five vs. Six & (b) Dogs vs. Cats & (c) COV. vs. Bact. \\
        \hline \hline
        w/o $L_s$ & ${63.88 \pm 4.97}$ & ${69.92 \pm 6.50}$ & ${52.68 \pm 4.80}$ \\ 
        \hline
        w/o $L_t$ & ${3.57 \pm 1.93}$ & ${88.28 \pm 0.98}$ & ${48.19 \pm 5.92}$ \\ 
        \hline
        Re. $L_s$ & ${56.87 \pm 1.26}$ & ${89.45 \pm 0.92}$ & ${52.68 \pm 16.04}$ \\ 
        \hline
        Re. $L_t$ & ${13.16 \pm 1.64}$ & ${87.07 \pm 1.61}$ & ${42.47 \pm 5.21}$ \\ 
        \hline
        Re. $L_s$ \& $L_t$ & ${19.08 \pm 10.37}$ & ${86.95 \pm 0.93}$ & ${48.59 \pm 5.58}$ \\
        \hline \hline
        Proposed & \boldmath{${75.18 \pm 3.52}$} & \boldmath{${91.80 \pm 0.44}$} & \boldmath{${76.50 \pm 3.38}$} \\
        \hline
    \end{tabular}
\end{table}

In summary, the ablation studies show that our loss function design choices played an essential role in alleviating the inherited texture bias problem observed in training data, and proved that our method could mitigate texture bias across multiple datasets with high performance by adopting an image generation strategy. Our ablation studies also demonstrate superior performances shown by the experiments with replacement as they have the same settings with previous style transfer methods \cite{gatys2016image,li2017demystifying}. Our method not only achieved higher quantitative performance against style transfer competitors, but also showed better visual quality (Fig. \ref{fig_classification_ablation}).

\subsection{Unsupervised Domain Adaptation}

\subsubsection{GTAV-Cityscapes}

\begin{figure}[!t]
	\centering
	\includegraphics[width=0.7\columnwidth]{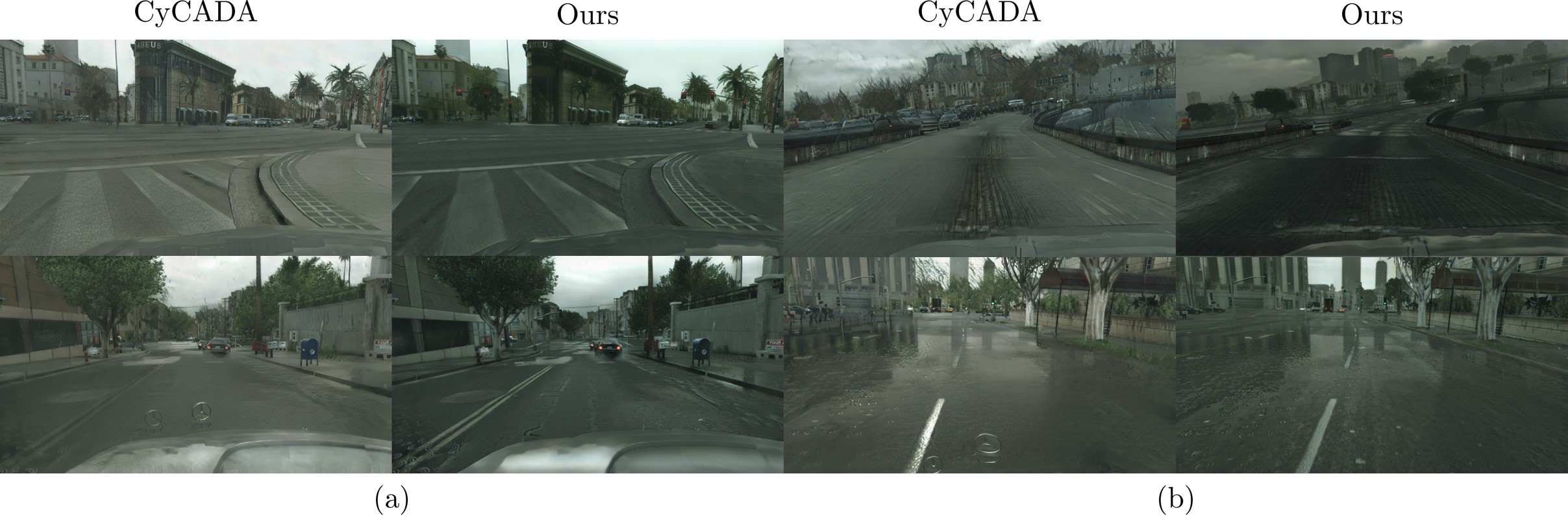}
    \caption{Qualitative results of CyCADA \cite{hoffman2018cycada} models and the proposed on the GTAV-Cityscapes dataset.}
	\label{fig_adaptation_gtav_cityscapes}
\end{figure}

In Table \ref{tab_segmentation_gtav_cityscapes}, we compare against state-of-the-art unsupervised domain adaptation methods. The proposed method outperforms all competitors when the translated images are employed for training by a state-of-the-art segmentation model. In addition, we qualitatively compare our method's translated images with CyCADA \cite{hoffman2018cycada} in Fig. \ref{fig_adaptation_gtav_cityscapes}. CyCADA employs cycle consistency loss to preserve content information; however, as we previously mentioned - content distortion in the translated image is observed when using this loss function alone. In first row of Fig. \ref{fig_adaptation_gtav_cityscapes}(a), CyCADA translates images appropriately without content distortions. However, in Fig. \ref{fig_adaptation_gtav_cityscapes}(b) extensive content distortions (\eg new cars are generated) with artifacts and noises are observed. Interestingly in second row of Fig. \ref{fig_adaptation_gtav_cityscapes}, CyCADA creates car emblems that are not present in GTAV samples, rather mostly observed in Cityscapes. In contrast, our method does not hallucinate car emblems or weird textures in the generated image.

\begin{table}[!t]
    \centering
    \footnotesize
    \caption{Unsupervised domain adaptation performance of state-of-the-art methods and the proposed model on GTAV-Cityscapes.}
    \label{tab_segmentation_gtav_cityscapes}
    \renewcommand{\tabcolsep}{1pt} 
    \begin{tabular}{l|lllllllllllllllllll|l}
        \hline
        & Road & S.W. & Bui. & Wall & Fen. & Pole & Tr.L. & Sign & Veg. & Ter. & Sky & Per. & Rid. & Car & Tru. & Bus & Tra. & M.B. & Bike & mIoU \\
        \hline
        CyCADA \cite{hoffman2018cycada} & 79.1 & 33.1 & 77.9 & 23.4 & 17.3 & 32.1 & 33.3 & 31.8 & 81.5 & 26.7 & 69.0 & 62.8 & 14.7 & 74.5 & 20.9 & 25.6 & 6.9 & 18.8 & 20.4 & 39.5 \\
        \who{Tsai}{tsai2018learning} & 86.5 & 36.0 & 79.9 & 23.4 & 23.3 & 23.9 & 35.2 & 14.8 & 83.4 & 33.3 & 75.6 & 58.5 & 27.6 & 73.7 & 32.5 & 35.4 & 3.9 & 30.1 & 28.1 & 42.4 \\
        \who{Li}{li2019bidirectional} & 91.0 & 44.7 & 84.2 & 34.6 & 27.6 & 30.2 & 36.0 & 36.0 & 85.0 & 43.6 & 83.0 & 58.6 & 31.6 & 83.3 & 35.3 & 49.7 & 3.3 & 28.8 & 35.6 & 48.5 \\
        CBST \cite{zou2018unsupervised} & 91.8 &  53.5 & 80.5 & 32.7 & 21.0 & 34.0 & 28.9 & 20.4 & 83.9 & 34.2 & 80.9 & 53.1 & 24.0 &  82.7 & 30.3 & 35.9 &  16.0 & 25.9 & 42.8 & 45.9 \\
        \who{Mei}{mei2020instance} & 94.1 & 58.8 & 85.4 & 39.7 & 29.2 & 25.1 & 43.1 & 34.2 & 84.8 & 34.6 & 88.7 & 62.7 & 30.3 & 87.6 & 42.3 & 50.3 & 24.7 & 35.2 & 40.2 & 52.2 \\
        \who{Kim}{kim2020learning} & 92.9 &  55.0 & 85.3 & 34.2 & 31.1 & 34.9 & 40.7 & 34.0 & 85.2 & 40.1 & 87.1 & 61.0 & 31.1 & 82.5 & 32.3 & 42.9 & 0.3 & 36.4 & 46.1 & 50.2 \\
        DACS \cite{tranheden2021dacs} & 89.9 &  39.7 & 87.9 & 30.7 & 39.5 & 38.5 &  46.4 & 52.8 &  88.0 & 44.0 & 88.8 & 67.2 & 35.8 & 84.5 & 45.7 & 50.2 &  0.0 & 27.3 & 34.0 & 52.1 \\
        \cite{xie2021segformer} + Src & 78.4 & 27.9 & 82.5 & 32.3 & 34.6 & 34.9 & 54.3 & 28.8 & 87.5 & 36.7 & 81.6 & 70.6 & 25.9 & 86.6 & 39.7 & 24.9 & 9.8 & 33.6 & 34.6 & 47.6 \\
        \hline
        Ours & 88.7 & 46.7 & 87.2 & 48.3 & 32.8 & 36.8 & 57.5 & 23.3 & 88.0 & 42.4 & 87.7 & 67.7 & 34.2 & 89.0 & 47.5 & 48.4 & 26.5 & 40.0 & 47.5 & \boldmath{$54.7$} \\
        \hline
    \end{tabular}
\end{table}

\begin{table}[!t]
    \centering
    \small
    \caption{Segmentation performance of image image translation models, and the proposed model on the prostate dataset. DSC with $\pm$ std was used as the metric.}
    \label{tab_segmentation_prostate}
    \begin{tabular}{l||c|c|c}
        \hline
        Method & A & B & Avg. \\
        \hline \hline
        CycleGAN \cite{zhu2017unpaired} & ${32.18 \pm 9.46}$ & ${35.54 \pm 10.58}$ & ${33.86}$ \\ 
        \hline
        UNIT \cite{liu2017unit} & ${37.43 \pm 10.20}$ & ${39.15 \pm 10.96}$ & ${38.29}$ \\ 
        \hline
        MUNIT \cite{huang2018munit} & ${26.62 \pm 9.21}$ & ${37.90 \pm 11.22}$ & ${32.26}$ \\ 
        \hline
        DRIT++ \cite{lee2020drit++} & ${33.71 \pm 12.17}$ & ${40.52 \pm 11.33}$ & ${37.12}$ \\ 
        \hline
        CUT \cite{park2020cut} & ${39.76 \pm 0.72}$ & ${32.11 \pm 9.91}$ & ${35.93}$ \\
        \hline \hline
        Proposed & \boldmath{${57.75 \pm 1.33}$} & \boldmath{${48.60 \pm 13.24}$} & \boldmath{${53.18}$} \\
        \hline
        \hline
    \end{tabular}
\end{table}

\begin{figure}[!t]
	\centering
	\includegraphics[width=0.5\columnwidth]{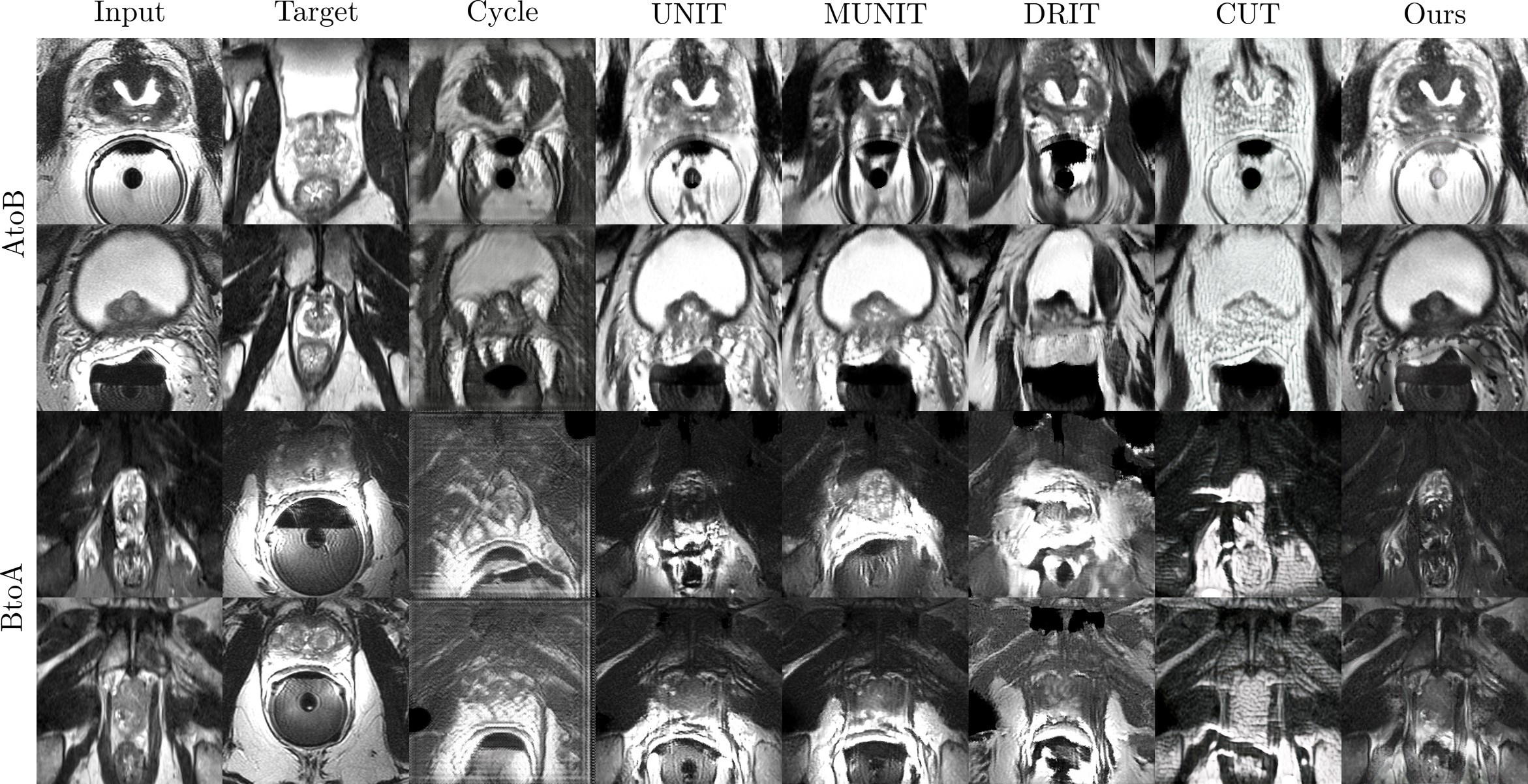}
    \caption{Qualitative results of five image translation models and the proposed on the prostate dataset.}
	\label{fig_adaptation_prostate}
\end{figure}

\subsubsection{Prostate}

We report the segmentation performance on prostate datasets in Table \ref{tab_segmentation_prostate}. To better highlight performance, we also present visual comparison on the quality of the generated images in Fig. \ref{fig_adaptation_prostate}. Overall, our method reports the best performance against existing methods. CycleGAN, DRIT, UNIT BtoA, and MUNIT showed extensive content distortions in the source image and while UNIT AtoB had relatively fewer content distortions with fewer texture updates. In addition, CUT translated images included artifacts (noise) and checkerboard patterns initially absent in the target image. Based on the reported quantitative (Table \ref{tab_segmentation_prostate}) and qualitative results (Fig. \ref{fig_adaptation_prostate}), we believe that the visual quality of the translated images is highly correlated to actual model performance. In contrast to prior approaches, our method shows high-quality image translation able to maintain the content of the source image while successfully transferring the texture of the target image using the proposed losses.


\section{Conclusion}\label{sec:conclusion}
In this work, we proposed to leverage an image translation framework to augment a dataset with new instances that mitigate or adapt to biases present in the data. We showed that transferring texture while maintaining content information is a valid choice for addressing data biases such as color and texture, as well as more complex and realistic biases induced by different scanning protocols, and/or images from graphics game engines. Extensive experiments and ablations support the choice of spatial- and texture-based losses for the accurate translation of texture without modifying content. This is a vital requirement in most medical and industrial domains that have strict structural constraints on their images, and cannot solely rely on web or social media to collect a large number of data samples. Usually, data collection is often limited and biased to some local settings (\eg acquisition equipment or protocols), where the labels might be aligned with features not related to the classes intrinsic properties. Notably, we empirically demonstrate the effectiveness of our method to solve a real world scenario by handling biases induced by acquisition protocols in the COVID-19 vs. Bacterial pneumonia and Prostate experiments, and we report high accuracy with large margins across all evaluated datasets; with higher quality generated images compared to recent methods.





\bibliographystyle{model5-names}\biboptions{authoryear}
\bibliography{paper_ref}







\end{document}